\documentclass[a4paper, 10pt, conference]{ieeeconf}
\usepackage{FG2027}

\FGfinalcopy

\IEEEoverridecommandlockouts
\usepackage{graphicx}
\usepackage{amsmath}
\usepackage{amssymb}
\usepackage{booktabs}
\usepackage{xcolor}
\usepackage{url}
\usepackage{capt-of}
\usepackage{placeins}
\graphicspath{{figures/}{./}}

\def\FGPaperID{237}

\title{\LARGE \bf
Faces in Things, Faces in Nothing:\\
Probing Face Selectivity in Vision Models with Pareidolia
}

\author{\parbox{16cm}{\centering
    {\large Qianpu Chen$^1$, Derya Soydaner$^1$, and Rob Saunders$^1$}\\
    {\normalsize
    $^1$ LIACS (Leiden Institute of Advanced Computer Science), Leiden University, Leiden, The Netherlands}}
    \thanks{Code will be released upon publication.}%
}

\begin{document}

\ifFGfinal
\thispagestyle{empty}
\pagestyle{empty}
\else
\author{Anonymous FG2027 submission\\ Paper ID \FGPaperID \\}
\pagestyle{plain}
\fi
\twocolumn[{%
\renewcommand\twocolumn[1][]{#1}%
\maketitle
\vspace{-1.5em}
\begin{center}
\includegraphics[width=\textwidth]{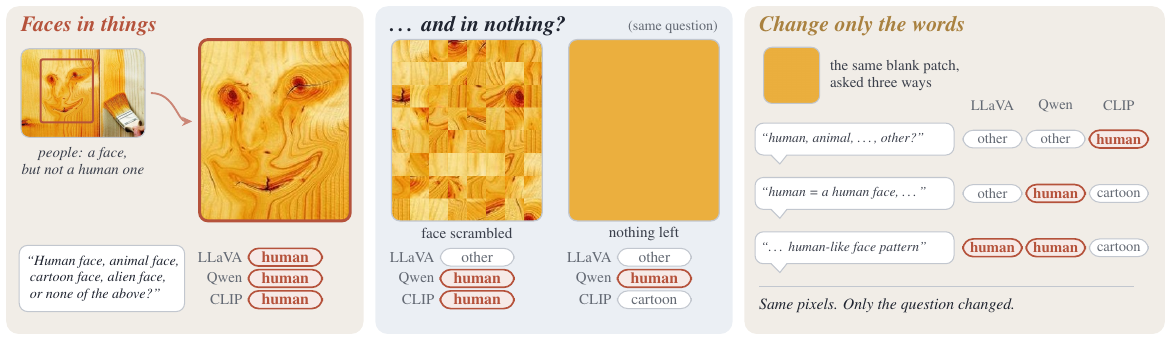}
\captionof{figure}{A face in wood, and a face in nothing. This is face pareidolia: people see a smiling face in this wood grain, yet know it is not a person's. Asked what kind of face it is, three vision models answer Human (left). Shuffle the tiles until the face is gone, and two of them still answer Human; flatten the region to a single color, and Qwen2-VL still does (middle). Keep the patch fixed and change only the question, and the answer changes with it (right). Calling something Human is not the same as seeing its face.}
\label{fig:pareidolia}
\end{center}
\vspace{0.6em}
}]

\begin{abstract}
An electrical outlet can look like a face without depicting a person. Face pareidolia is thus a revealing probe of how vision models turn ambiguous visual evidence into category judgments. When a model calls such an object \emph{Human}, does that answer depend on its face-like arrangement? Using the FacesInThings dataset, we test six vision models, from a vision-only ViT to CLIP and generative vision-language models, with face detectors as a reference. Each face-like region is paired with three controls from the same photograph that lack the arrangement: a background crop, the region with its tiles shuffled, and its average color. A Human answer driven by the arrangement should appear less often on these controls. We also vary how the categories are described. Three findings emerge. First, models respond to illusory faces, but not as people do: face detectors rarely return a face for these regions, yet CLIP ViT-L/14 and two generative models label roughly half of them Human when the categories are described as kinds of face. Annotators, by contrast, read these faces as non-human, such as animal or cartoon faces. For these models, a face is a human face by default. Second, similar Human rates hide opposite behavior. LLaVA-1.5 answers Human less often once the arrangement is removed; Qwen2-VL calls every average-color patch Human and describes a person in each; CLIP ViT-B/32 calls shuffled regions Human more often than real faces; SmolVLM rarely answers Human, but mostly for face-like regions. Third, rewording the categories changes whether Human answers favor the face at all, removing this preference in LLaVA-1.5 and creating it in CLIP. A Human answer is not evidence of a face: vision models can keep giving it after the face is gone.
\end{abstract}

\section{INTRODUCTION}

When an electrical outlet or a grain of wood seems to look back at us, we see a face and still know that no one is there (Fig.~\ref{fig:pareidolia})~\cite{liu2014seeing,wardle2020rapid,caruana2022objects}. This is face pareidolia: people keep the face-like arrangement separate from the judgment of what kind of being it belongs to, and read most illusory faces as animal, cartoon, or otherwise non-human~\cite{hamilton2024seeing}. When a vision model calls such an object \emph{Human}, is that answer about the face? The question matters, since whether something is a person shapes how systems handle privacy, safety, and social interaction, and ambiguous faces in shadow, reflection, artwork, or clutter are common in practice.

The answer alone cannot tell. Vision-language models (VLMs) are mostly evaluated by whether they assign the correct label to an image, but a correct label does not reveal which aspects of the image support the judgment: an answer may rest on the visual evidence, on the wording of the categories, or on what the model tends to say regardless of input. Telling these apart requires asking how the answer changes when specific visual content is removed or replaced. Pareidolia makes this especially pointed: its faces are ambiguous by nature, and when the visual evidence is weak, wording and answer tendencies have the most room to shape the answer.

Research on face perception offers a way to test whether a response is sensitive to facial structure.
A response is called face-selective when it is stronger for intact faces than for the same images with their spatial arrangement scrambled or removed~\cite{kanwisher1997ffa,tsao2006cortical}. Pareidolia is a natural stimulus for this test: a face-like region contains the arrangement of a face but no face, and how people read it is known. An answer grounded in the face-like arrangement should be given more often to the region than to the same content once the arrangement is gone.

We pose the same five-way question to six model interfaces: a vision-only ViT with label prototypes, CLIP ViT-B/32 and ViT-L/14 with category texts, and three generative VLMs (LLaVA-1.5, Qwen2-VL, SmolVLM). Each face-like region is paired with three controls from the same photograph, a background crop, the region scrambled into $8{\times}8$ tiles, and a mean-color patch, and we vary only how the five categories are described. Face detectors serve as a reference for whether the regions register as faces at all.

Face detectors rarely accept these regions as faces. Once models must choose among categories described as kinds of face, however, LLaVA-1.5 and Qwen2-VL label roughly half of the non-human-annotated regions Human, as if a face were a human face by default, yet their responses to controls differ sharply: LLaVA-1.5 answers Human less often once the arrangement is removed, whereas Qwen2-VL calls every mean-color patch Human, and CLIP ViT-B/32 calls scrambled regions Human more often than real faces. In this sense, Qwen2-VL and CLIP find faces in nothing. Changing only the descriptions changes this preference itself, removing it in LLaVA-1.5 and creating it in CLIP.

Our contributions are fourfold:
\begin{itemize}
    \item Inspired by methods used to study face selectivity in neuroscience, 
    we introduce \emph{pareidolic selectivity}, a framework and measure that tests whether a vision model's response is about a face. It applies to any vision model, open or closed; we test six models, from a vision-only ViT to generative VLMs, and release the code.
    \item We reveal that, for several vision models, a face is a human face by default: they respond to illusory faces but label about half of those that annotators judged non-human as Human, even when, as in LLaVA-1.5, the answer follows the face.
    \item We show that Human answers can also outlive the face: in four of six models they persist or grow once the face-like arrangement is removed, and on unchanged images, describing the categories as kinds of face amplifies them.
    \item We demonstrate that face selectivity is not a fixed property of a model but emerges from the model and the question together. More broadly, on ambiguous inputs a label is evidence of perception only when read against controls and the question that produced it.
\end{itemize}

\section{RELATED WORK}

\paragraph{Face selectivity and pareidolia in human vision} Whether a neural response is about faces has long been tested by contrast: face-selective regions and cells respond more strongly to intact faces than to objects or scrambled faces that keep low-level content but lose the facial arrangement~\cite{kanwisher1997ffa,tsao2006cortical}. Pareidolia probes the same machinery from the other side. Faces reported in pure noise engage face-selective cortex~\cite{liu2014seeing}, illusory faces in objects evoke face-like neural responses within a few hundred milliseconds~\cite{wardle2020rapid}, and they reach awareness faster than matched objects~\cite{caruana2022objects}, while the identity of the object remains available. Our test borrows this logic and asks whether a model's answer goes with the arrangement when the arrangement is removed.

\paragraph{Faces and pareidolia in artificial vision}
Deep networks develop face-selective units, whether trained for object or face recognition~\cite{dobs2022brain} or untrained~\cite{baek2021untrained}, with selectivity defined, as in neuroscience, by contrasting faces with non-face images. For pareidolia, Hamilton et al.~\cite{hamilton2024seeing} introduced the FacesInThings dataset and showed that face detectors respond to illusory faces far less often than people do. Gupta and Dobs~\cite{gupta2025humanlike} found that a network trained for face identification and object categorization relies on face-like features when classifying pareidolic stimuli as faces. These studies ask whether internal units or detectors respond to faces; we ask whether a model's verbal answer, calling a face-like object Human, is face-selective, and how that changes with the language that defines the categories.

\paragraph{Grounding and language priors in VLMs}
A recurring concern is that VLMs answer from language rather than from the image. Many benchmark questions can be answered without the image~\cite{chen2024mmstar}; models fall back on language priors when visual evidence is weak~\cite{leng2024vcd,favero2024m3id,lee2025vlind}, tend to affirm that objects are present~\cite{li2023pope}, answer visual illusions from prior knowledge~\cite{guan2024hallusionbench}, and agree with human illusory percepts only in part~\cite{zhang2023grounding}. Prompt wording shifts which reading of a bistable image is reported~\cite{panagopoulou2024bistable}, and CLIP's zero-shot predictions depend on how category texts are written~\cite{zhou2022coop}. These studies score answers on target images against a reference or ask whether answers change with wording. We ask instead what an answer depends on, which requires controls: the same question is asked of a face-like region and of controls that remove its arrangement, while the description is varied with the images fixed.

\section{METHOD}
\label{sec:method}

\begin{figure*}[t]
    \centering
    \includegraphics[width=\linewidth]{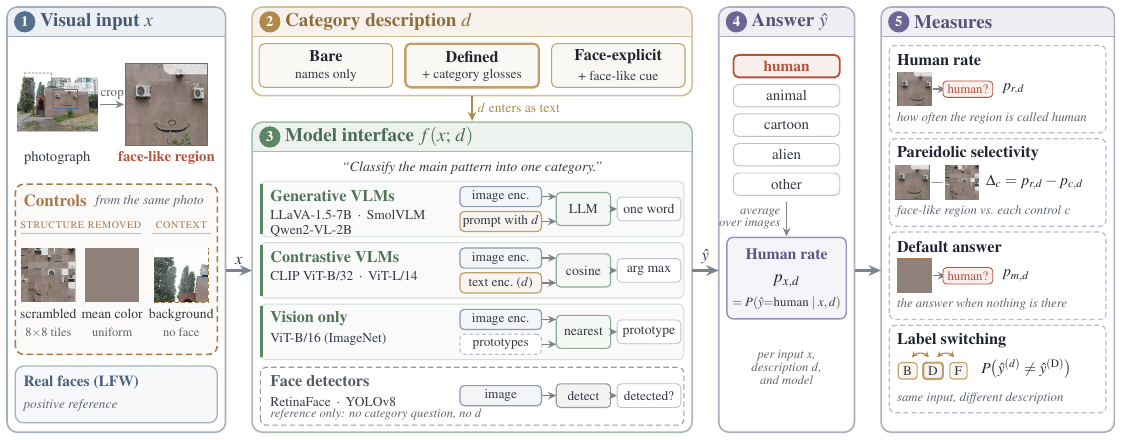}
    \caption{The pareidolic selectivity test. (1) Each face-like region is paired with three controls from the same photograph: a background crop (dashed box) and scrambled and mean-color versions of the region. (2) The five categories are described in one of three ways (Table~\ref{tab:prompts}). (3) Models receive the description through a prompt, through category texts, or not at all; face detectors show whether the regions register as faces, and real-face photographs (LFW) give a positive reference: how often actual faces are called Human. (4) Averaging Human answers over images gives a Human rate $p_{x,d}$. (5) Comparing rates yields pareidolic selectivity (region vs.\ control), the default answer (mean-color patch), and label switching (across descriptions).}
    \label{fig:framework}
\end{figure*}

Our framework, pareidolic selectivity, asks whether a model's Human answer depends on the face-like arrangement of a region (Fig.~\ref{fig:framework}). Following intact-versus-scrambled comparisons used in face perception research, it keeps the photograph fixed and varies two factors: the input, either a face-like region or one of three controls from the same photograph that lack its arrangement, and the description of the five categories. Each input $x$ is shown to a model with a description $d$, and the model returns one category $\hat y$.

\subsection{Dataset}
\label{sec:task}
\begin{figure*}[t]
    \centering
    \includegraphics[width=0.95\linewidth]{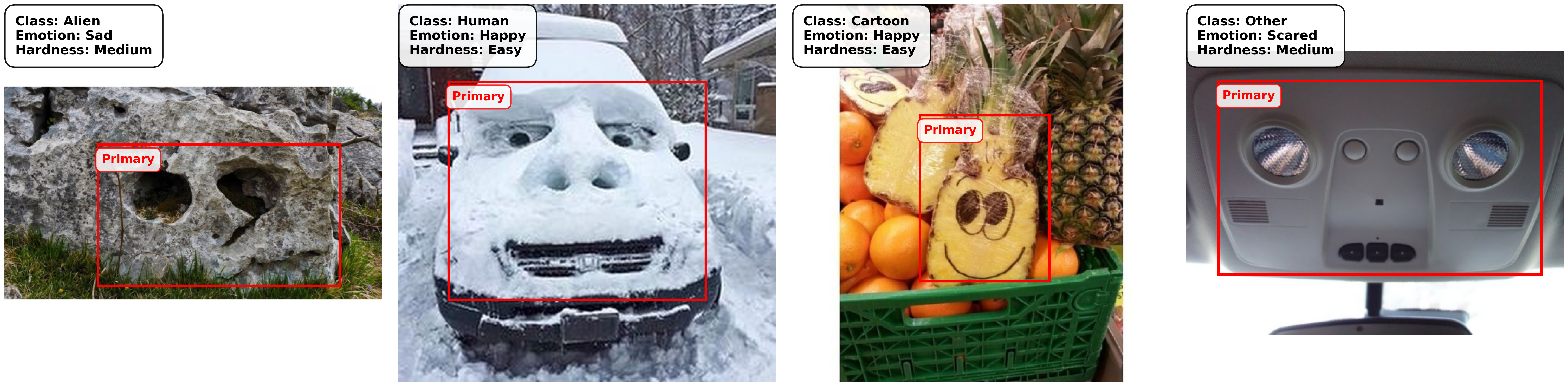}
    \caption{Examples from FacesInThings~\cite{hamilton2024seeing}. Every face is illusory. Labels give how annotators read each face (human-like, animal, cartoon, alien, or other), not what the object is. Boxes mark the annotated regions from which crops are taken.}
    \label{fig:dataset}
\end{figure*}

FacesInThings~\cite{hamilton2024seeing} is a collection of photographs of everyday objects and scenes in which people see a face (Fig.~\ref{fig:dataset}). Its annotations describe how people read each illusory face, not what the object is: a face may read as human-like, as an animal, a cartoon character, an alien or monster, or as none of these, and the boundaries between these readings are soft. We use $4{,}536$ images. From each image we take the region marked as primary, or the first annotated region when none is marked, and assign it one of five categories: Human, Animal, Cartoon, Alien, or Other, from the annotators' judgment of what kind of being the face belongs to; $14$ regions without this annotation are excluded (Supp.~S1). A Human region is thus an illusory face that people read as human-like, never a photograph of a person.

Of the 4,522 labeled regions, 1,114 are read as human-like; we call them Human-annotated. The other 3,408 are read as Animal (1,035), Cartoon (700), Alien (335), or Other (1,338); we call them non-human. Our main analyses use the non-human regions: people read these faces as something other than human, so a Human answer there marks a model adopting a human reading that people did not. The Human-annotated regions serve as a point of comparison. Comparisons with controls use $400$ of the non-human regions, taken in a fixed order from images in which a background crop can be placed away from every annotated face.

\subsection{Inputs and Controls}
\label{sec:controls}
Following the face-selectivity contrast~\cite{kanwisher1997ffa}, we pair each face-like region with three controls from the same photograph that lack its face-like arrangement (Fig.~\ref{fig:framework}, left).

The \emph{scrambled} control cuts the region into an $8{\times}8$ grid and shuffles the tiles, keeping its colors and local texture but breaking the arrangement of its features. The \emph{mean-color} control replaces the region with a patch of its average color, removing all structure. The \emph{background} control is a square crop from elsewhere in the same photograph, placed outside every annotated face. It shares the lighting, camera, and scene of the region but contains no face-like arrangement.

As a positive reference, models also see photographs of real faces drawn at random from Labeled Faces in the Wild (LFW)~\cite{huang2008lfw}, in the tightly cropped format of its standard distribution: $400$ for the generative models and $500$ for CLIP ViT-B/32. CLIP's structure-removing controls are additionally run on all $4{,}536$ regions (Supp.~S11).

\subsection{Category Descriptions}
\label{sec:descriptions}
Category names alone leave open what each category includes, and the description given to a model fills that in. We use three descriptions that share the five names (Table~\ref{tab:prompts}): \emph{Bare} gives the names alone; \emph{Defined}, our reference, glosses each category as a kind of face; \emph{Face-explicit} adds a face-like cue and widens Human and Animal to face-like patterns. The same glosses serve as prompt text for the generative models and as category texts for CLIP, so both families see matching descriptions while images and models stay fixed. For CLIP, we also use an ensemble of four texts per category, whose Other texts describe concrete non-face content (Supp.~S3).

\begin{table}[t]
\centering
\caption{Category descriptions. Defined is shown in full; the other descriptions are listed as changes to it. All use the same five category names. Generative models read the glosses in a prompt that asks for one category name; CLIP uses each gloss as a category text. Full prompts and texts are in Supp.~S3.}
\label{tab:prompts}
\setlength{\tabcolsep}{4pt}
{\footnotesize
\begin{tabular}{@{}l p{5.9cm}@{}}
\toprule
\multicolumn{2}{@{}l}{\textit{Defined} (reference)}\\
\quad \texttt{human}   & a human face \\
\quad \texttt{animal}  & an animal face \\
\quad \texttt{cartoon} & a cartoon / emoji / animated character face \\
\quad \texttt{alien}   & an alien / monster / unrealistic creature face \\
\quad \texttt{other}   & none of the above \\
\midrule
\textit{Bare}          & glosses removed; names only \\
\addlinespace
\textit{Face-explicit} & adds ``This crop may contain a face-like pattern.'' (not used for CLIP); \texttt{human} and \texttt{animal} gain ``\ldots\ or human-like / animal-like face pattern'' \\
\addlinespace
\textit{CLIP ensemble} & four texts per category; \texttt{other} described by non-face content, e.g., ``a texture or pattern that is not a face'' \\
\bottomrule
\end{tabular}
}
\end{table}

\subsection{Models}
\label{sec:interfaces}
All models see the same inputs and choose among the same five categories, but they take in the description in different ways (Fig.~\ref{fig:framework}, middle).

\textbf{Generative VLMs.} LLaVA-1.5-7B~\cite{liu2023llava,liu2024improved}, Qwen2-VL-2B~\cite{wang2024qwen2vl}, and SmolVLM~\cite{marafioti2025smolvlm} receive the input with a question that lists the five categories and their glosses and asks for one category name. Answers are decoded greedily; checkpoints and settings are listed in Supp.~S2.

\textbf{Contrastive models.} CLIP ViT-B/32 and ViT-L/14~\cite{radford2021clip} compare the input with one text per category, taken from Table~\ref{tab:prompts}, and return the closest category.

\textbf{Vision only.} An ImageNet-pretrained~\cite{russakovsky2015imagenet} ViT-B/16~\cite{dosovitskiy2020image} receives no text. Each category is represented by the average feature of all other annotated regions in that category, and an input is assigned to the nearest one; leaving the evaluated region out of its own category keeps its label from entering the prediction. This model shows how often visual similarity alone leads to Human.

\textbf{Face detectors.} As a reference for whether the regions register as faces at all, we run RetinaFace~\cite{deng2020retinaface} and YOLOv8~\cite{jocher2023yolov8} on each annotated region and record whether RetinaFace returns a face and whether YOLOv8 returns a person.

\subsection{Measures}
\label{sec:measures}
For every model, description $d$, and input type $x$, we measure how often the answer is Human (Fig.~\ref{fig:framework}, right),
\begin{equation}
p_{x,d} = P(\hat y = \mathrm{Human} \mid x, d),
\end{equation}
where a Human answer is the name \texttt{human} for the generative models, the Human text for CLIP, and the Human prototype for ViT. Under Defined, Human is glossed as ``a human face'', so a Human answer concerns how the input is read, not whether a person is present. On face-like regions ($x{=}r$), we call this the \emph{Human rate}. Three further measures compare these rates.

\emph{Pareidolic selectivity}, a target--control difference in Human answers, asks whether Human goes with the face-like arrangement. For a control $c$ it is the paired difference
\begin{equation}
\Delta_c = p_{r,d} - p_{c,d},
\end{equation}
computed on the same images. A positive $\Delta_c$ means Human is given mainly when the arrangement is present. Because $\Delta_c$ is bounded by how often a model answers Human at all, we read selectivity from its sign and confidence interval and report Human rates alongside it. The \emph{default answer} $p_{m,d}$ is the Human rate on mean-color patches: what a model says when there is nothing to see. \emph{Label switching} is the fraction of regions whose answer changes when the description changes, which captures shifts that the Human rate alone can hide.

Confidence intervals are $95\%$ percentile bootstraps over images ($2{,}000$ resamples), resampling all inputs and descriptions of an image together so that paired differences keep their pairing. To see what the generative models report without the category task, we also ask them to describe $20$ mean-color and $20$ scrambled inputs in a few words. To test whether Qwen2-VL's answers depend on the image at all, we also give it the same prompts with no image, with human placed first or last in the category list. Three SmolVLM answers outside the five categories are excluded (Supp.~S5).

\section{RESULTS}

We answer three questions in turn. Do models respond to illusory faces, and do they read them as people do (Sec.~\ref{sec:res_readings})? Are Human answers given more often to the regions than to controls that lack the face-like arrangement (Sec.~\ref{sec:res_select})? Does this preference change when only the category description changes (Sec.~\ref{sec:res_desc})? Unless stated otherwise, rates refer to regions whose faces annotators judged non-human, comparisons with controls use the $400$-image subset, and Defined is the reference description.

\subsection{Models Respond to Illusory Faces, but Not as People Do}
\label{sec:res_readings}

\begin{figure*}[t]
    \centering
    \includegraphics[width=0.86\linewidth]{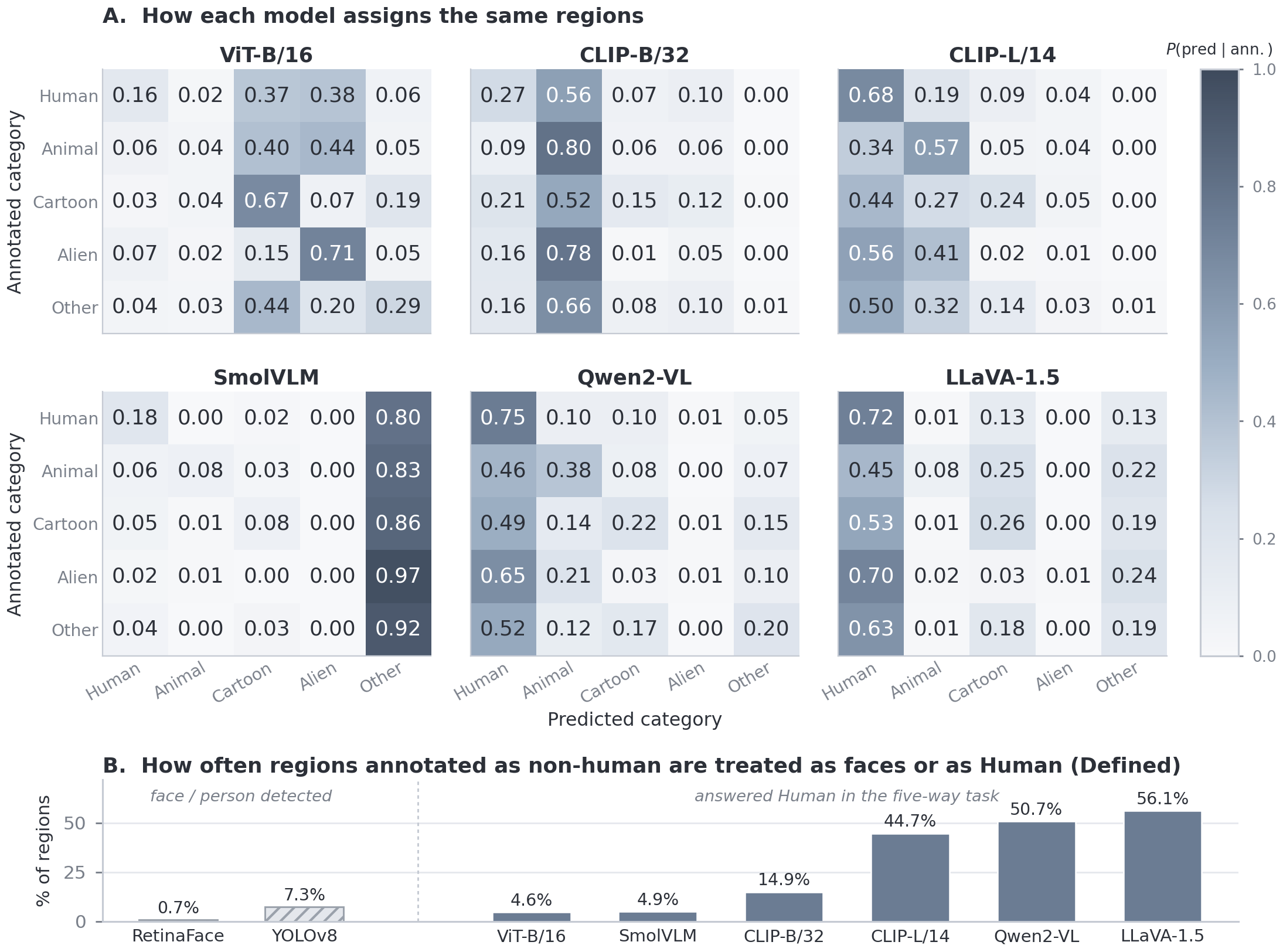}
    \caption{Readings of the same face-like regions under the Defined description. \textbf{A:} Distribution of predicted categories within each annotated category (rows sum to one). \textbf{B:} On the $3{,}408$ non-human-annotated regions, how often face detectors return a face or person (hatched) and how often each model answers Human.}
    \label{fig:profiles}
\end{figure*}

Every region in FacesInThings is an illusory face, and face detectors, which apply a real-face criterion, rarely accept one. RetinaFace returns a face for about $1\%$ of the regions, more often on regions annotated as human-like ($2.5\%$ against $0.7\%$) or as easy to spot (about seven times more often than for hard ones). YOLOv8 reports a person for about $9\%$ of regions, but as often for background crops from the same images ($6.5\%$ against $6.25\%$), so its responses are not tied to the face (Supp.~S8).

The six models instead answer a five-way question whose categories are those of the dataset's annotation (Sec.~\ref{sec:task}). Most of them respond to these regions as faces: LLaVA-1.5, Qwen2-VL, and ViT, which can all answer Other, assign a kind of face to $81$--$88\%$ of the regions, where RetinaFace accepts about $1\%$. SmolVLM is the exception and places nearly nine in ten in Other. The models also follow how the face reads to people: every model answers Human more often for regions annotated as human-like than for the rest ($72\%$ against $56\%$ for LLaVA-1.5; Supp.~S9). Where people tell animal, cartoon, and alien faces apart, however, several models converge on Human (Fig.~\ref{fig:profiles}). Of the regions annotators read as non-human, ViT and SmolVLM call fewer than $5\%$ Human, CLIP ViT-B/32 about $15\%$, and CLIP ViT-L/14, Qwen2-VL, and LLaVA-1.5 between $45\%$ and $56\%$.

Similar Human rates do not imply similar readings, and each model has a characteristic way of resolving the ambiguity (Fig.~\ref{fig:profiles}A). SmolVLM is conservative: it places nearly nine in ten regions in Other and commits to a kind of face only rarely, although it calls almost every real face Human, so its low rate does not reflect an inability to use the category. ViT, which relies only on visual similarity to annotated examples, assigns most regions to a kind of face, but three quarters to non-human kinds, cartoon or alien. CLIP ViT-B/32 settles mostly on Animal, whereas the larger CLIP ViT-L/14 shifts toward Human. LLaVA-1.5 and Qwen2-VL resolve the ambiguity toward Human across every annotated category, including faces that people read as animals. For these models, a face is a human face by default.

These profiles show which labels models assign, not whether Human answers favor the face-like regions; we turn to that next.

\subsection{Human Answers Do Not Consistently Favor the Face-like Region}
\label{sec:res_select}

\begin{table*}[t]
\centering
\caption{Human rates (\%) for face-like regions (R) and their controls on the same $400$ images: background (Bg), scrambled (Scr), and mean color (M). Bold marks conditions in which the region is called Human significantly more often than every control (all three $\Delta_c>0$, $95\%$ CIs excluding zero; all $\Delta_c$ with CIs in Supp.~S6).}
\label{tab:matrix}
\setlength{\tabcolsep}{3.6pt}
{\footnotesize
\begin{tabular}{l cccc c cccc c cccc}
\toprule
 & \multicolumn{4}{c}{Bare} && \multicolumn{4}{c}{Defined} && \multicolumn{4}{c}{Face-explicit} \\
\cmidrule{2-5}\cmidrule{7-10}\cmidrule{12-15}
Model & R & Bg & Scr & M && R & Bg & Scr & M && R & Bg & Scr & M \\
\midrule
LLaVA-1.5-7B & $11.0$ & $16.8$ & $5.8$ & $12.0$ && $\mathbf{59.0}$ & $12.2$ & $25.0$ & $0.0$ && $96.2$ & $99.5$ & $100$ & $100$ \\
Qwen2-VL-2B & $8.8$ & $8.5$ & $10.5$ & $0.0$ && $51.2$ & $51.5$ & $59.0$ & $100$ && $80.5$ & $86.0$ & $98.8$ & $100$ \\
SmolVLM & $\mathbf{4.0}$ & $1.5^\ddagger$ & $0.2$ & $0.0$ && $\mathbf{4.8}$ & $0.2$ & $0.2$ & $0.0$ && $\mathbf{6.0}$ & $1.2$ & $0.5$ & $0.0$ \\
\addlinespace
CLIP ViT-B/32 & $22.5$ & $14.5$ & $41.2$ & $43.2$ && $18.8$ & $28.2$ & $86.0$ & $31.0$ && $7.8$ & $14.5$ & $74.8$ & $3.5$ \\
CLIP ViT-L/14 & $58.8$ & $24.5$ & $67.8$ & $87.8$ && $43.8$ & $50.5$ & $86.8$ & $86.8$ && $16.5$ & $10.5$ & $81.5$ & $27.8$ \\
CLIP ViT-B/32$^\dagger$ & & & & && $\mathbf{36.2}$ & $29.8$ & $22.0$ & $0.0$ && & & & \\
CLIP ViT-L/14$^\dagger$ & & & & && $\mathbf{35.5}$ & $19.5$ & $19.8$ & $0.0$ && & & & \\
\addlinespace
ViT-B/16 & \multicolumn{4}{c}{} && $5.0$ & $9.0$ & $8.5$ & $0.5$ && \multicolumn{4}{c}{} \\
\bottomrule
\end{tabular}
}
\par\vspace{3pt}{\footnotesize ViT receives no description. $^\dagger$Four-text category ensemble. $^\ddagger$$397$ valid background answers.\par}

\end{table*}

\begin{figure*}[t]
    \centering
    \includegraphics[width=0.9\linewidth]{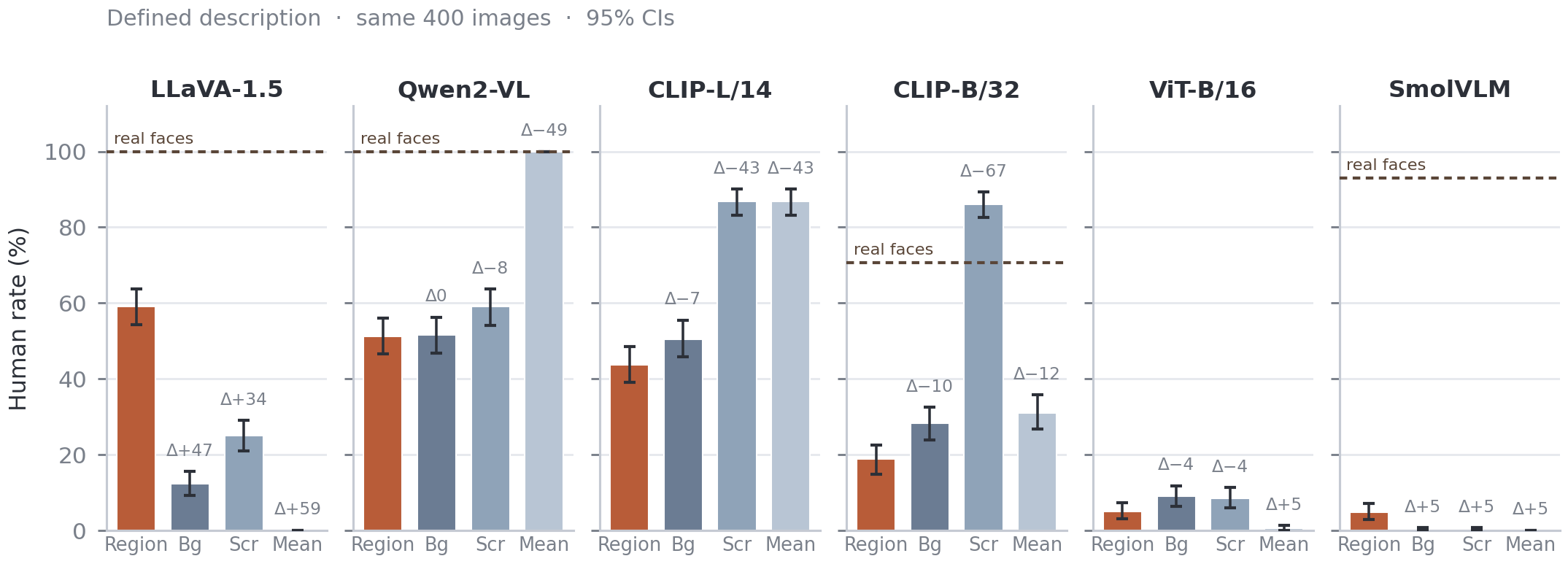}
    \caption{Human rates for face-like regions and their controls under Defined ($400$ images). A selective answer is highest for the region and falls once the arrangement is removed, as for LLaVA-1.5 (high rate) and SmolVLM (low rate). Qwen2-VL gives Human to backgrounds as often as to regions and to every mean-color patch; both CLIP models give it most often to scrambled regions. Dashed lines: real faces (LFW; not available for CLIP ViT-L/14 and ViT).}
    \label{fig:fourinput}
\end{figure*}

Fig.~\ref{fig:fourinput} and Table~\ref{tab:matrix} show, under the Defined description, how often each model calls a region and its three controls Human. Three contrasts organize the results.

LLaVA-1.5 and Qwen2-VL assign Human to roughly half of the target regions, but their control responses separate them sharply. LLaVA-1.5's Human answers fall step by step as the face-like arrangement is weakened: from every real face, to $59\%$ of regions, to a quarter of scrambled regions, an eighth of backgrounds, and no mean-color patch. Qwen2-VL gives backgrounds the same Human rate as the regions ($51\%$), gives scrambled regions slightly more, and calls every mean-color patch Human, exactly as it calls every real face. Asked simply to describe a mean-color patch, it describes a person who is not there in all $20$ cases (``a person standing in a room with a white wall''), where LLaVA-1.5 describes a wall and SmolVLM a solid color. We further tested where these answers come from by giving Qwen2-VL the same prompts with no image at all. It still answers Human under Defined, also when human is moved to the end of the category list, and describes the same person in the same words (Supp.~S7): for this model, a featureless patch is equivalent to no image. Judged by their target rates alone, LLaVA-1.5 and Qwen2-VL would be indistinguishable.

For CLIP, breaking the arrangement makes Human more likely, not less. Both CLIP models call scrambled regions Human far more often than intact ones ($86\%$ against $19\%$ for ViT-B/32 and $44\%$ for ViT-L/14), and ViT-B/32 calls scrambled regions Human more often than photographs of real faces, on the controlled subset and on all $4{,}536$ regions alike (Supp.~S11). CLIP recognizes most real faces but does not tie its Human answer to the face-like arrangement. Its classification rule helps explain why: it returns whichever category text matches the input best, with no threshold of facial evidence, so a scrambled region is called Human whenever it matches the Human text better than the four alternatives. Local face-like fragments may tip that comparison toward Human, in line with the reliance on local texture reported for other vision networks~\cite{geirhos2019texture}, and changing the alternatives reorganizes these answers (Sec.~\ref{sec:res_desc}).

SmolVLM and ViT both rarely answer Human, but a low rate does not imply the same selectivity. SmolVLM's few Human answers fall almost entirely on intact regions, giving a small but reliable preference over every control under every description, and it calls nearly all real faces Human: it keeps illusory faces apart from real ones, much as annotators do. ViT's equally rare Human answers are, if anything, slightly more frequent on backgrounds and scrambled regions than on the regions themselves.

These patterns do not stem from the controls: Qwen2-VL calls textured backgrounds Human more often than plain ones, and removing backgrounds in which a detector finds a person changes no result (Supp.~S4).

Under the reference description, the six models thus fall into three patterns. In LLaVA-1.5 and SmolVLM, Human follows the face-like arrangement, at opposite ends of the Human rate. In Qwen2-VL and ViT, Human is no more frequent on the regions than on backgrounds or scrambled regions, and for Qwen2-VL it is the default answer when nothing is there. In both CLIP models the relationship is inverted, with Human most frequent once the arrangement is broken.

\subsection{The Description Can Create or Remove Selectivity}
\label{sec:res_desc}

\begin{figure}[t]
    \centering
    \includegraphics[width=\linewidth]{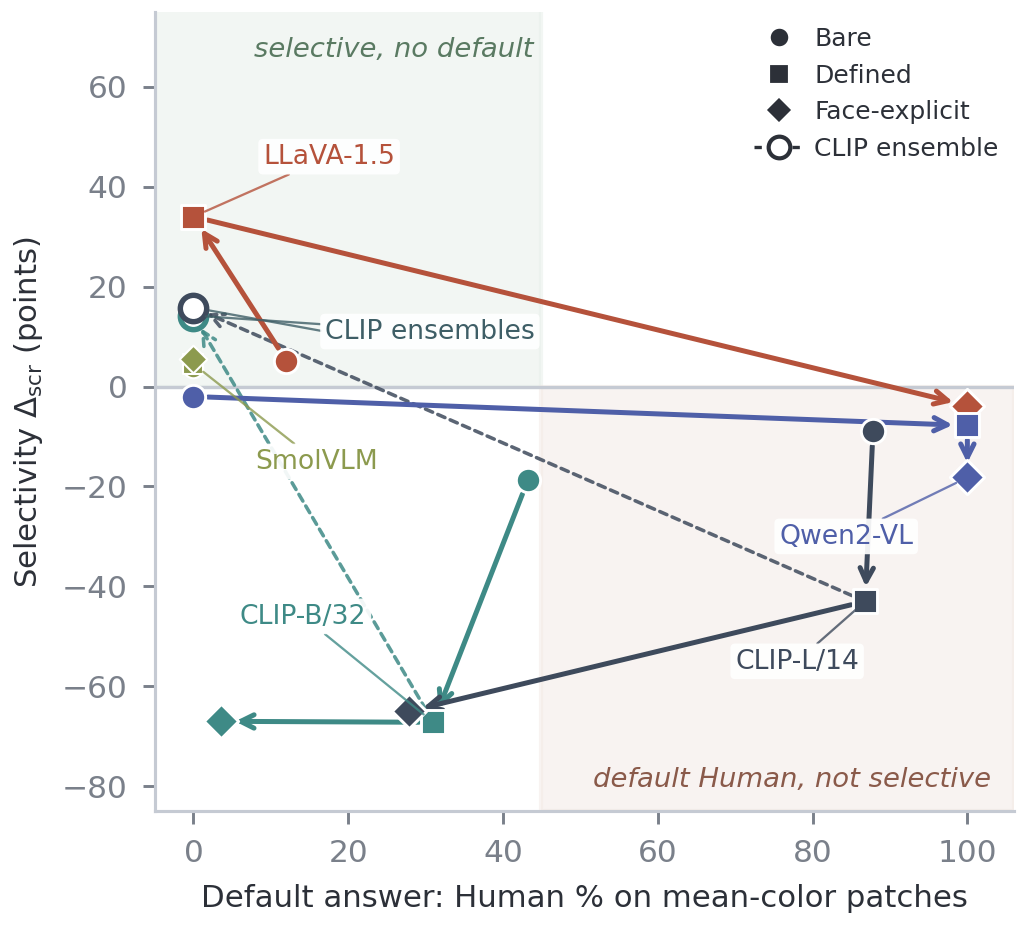}
    \caption{Changing only the category description moves models across the selectivity map. Arrows follow each model from Bare through Defined to Face-explicit ($400$ images). Upper left: Human follows the face-like arrangement and is not given to blank patches; lower right: Human is the default answer. Face-explicit moves LLaVA-1.5 from upper left to lower right; the four-text ensemble moves CLIP the other way. SmolVLM's points nearly coincide; ViT is omitted (Table~\ref{tab:matrix}).}
    \label{fig:map}
\end{figure}

A pareidolic face leaves its category open, and the description can fill in what the image does not settle. Holding images and models fixed, we now vary only the category description (Fig.~\ref{fig:map}; Table~\ref{tab:matrix}).

For the generative models, describing the categories as kinds of face is enough to make Human common. With category names alone, LLaVA-1.5 and Qwen2-VL call about one in ten non-human regions Human; once each category is glossed as a kind of face, about half; under Face-explicit, nearly all for LLaVA-1.5 and four in five for Qwen2-VL. SmolVLM barely moves. Human wins only once the categories are described as faces: among the kinds of face on offer, these models settle on the human one.

The description also changes what these answers depend on. LLaVA-1.5 shows no preference for the face-like region with names alone and a strong one with glosses. Under Face-explicit, which adds a face-like cue and widens Human to human-like face patterns, it calls almost every region, background, scrambled region, and mean-color patch Human, and the preference vanishes. Even the answer to an empty input is set by the description: Qwen2-VL calls no mean-color patch Human with names alone, but every one once glosses are added. Without an image, it picks whichever category is listed last under Bare, but Human under Defined wherever human is listed (Supp.~S7), so the glosses decide what its default answer is.

These shifts reach beyond Human: relative to Defined, names alone change the answer on about half or more of the regions for four models, and on one in five even for SmolVLM (Supp.~S10).

\begin{figure}[t]
    \centering
    \includegraphics[width=\linewidth]{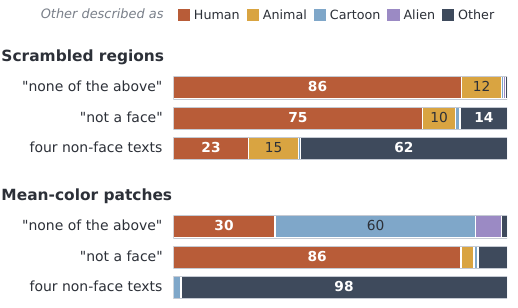}
    \caption{Where CLIP ViT-B/32 assigns inputs that lack the face-like arrangement (all $4{,}536$ regions). Rows are labeled by how the category texts describe Other: Defined, a minimal face-only set, and the four-text ensemble (Supp.~S3). Without a concrete alternative, these inputs fall to the face categories, mostly Human; with one, most go to Other.}
    \label{fig:dest}
\end{figure}

CLIP moves in the opposite direction. With one text per category, both CLIP models call scrambled regions Human more often than intact ones under all three descriptions. A four-text category ensemble reverses this relationship, producing higher Human rates for intact regions than for every control. The label destinations show what changes (Fig.~\ref{fig:dest}). Across all $4{,}536$ regions, single-text CLIP ViT-B/32 assigns Other to only $0.5\%$ of scrambled inputs, whereas the ensemble assigns it to $62\%$ of scrambled inputs and $98\%$ of mean-color patches; when Other is described only by negation (``not a face''), most mean-color patches become Human again (Supp.~S11). The reversal is thus a redistribution of control inputs from Human to Other: the category texts set the alternatives against which each image is classified, and an Other phrased as ``none of the above'' or ``not a face'' gives the image nothing to match.

The elevated Human response is not specific to tile scrambling. Under the single-text Defined description, phase-scrambled and strongly blurred versions of the regions, which contain no tile boundaries, receive Human answers in $84$--$98\%$ of cases for both CLIP models, while the ensemble sends most blurred and tile-scrambled inputs to Other (Supp.~S11). Together with LLaVA-1.5, these results show changes in both directions: the Face-explicit description removes LLaVA-1.5's preference for the face-like region, while the ensemble establishes one in CLIP.

Category descriptions thus change not only the prevalence of Human answers but whether those answers favor the face-like region over its controls.

\section{DISCUSSION}

\paragraph{For these models, a face is a human face}
Where people read an animal, a cartoon, or nothing human, several models settle on Human, even when the answer follows the face: LLaVA-1.5's Human rate falls from $59\%$ on the regions to none on mean-color patches, yet it calls nearly half of the faces that people read as animals Human (Supp.~S9). In people, registering a face and deciding whose face it is are separate steps: illusory faces evoke face-like responses while the object's identity stays available~\cite{wardle2020rapid,caruana2022objects}. We attribute the merge in models to how they learn the word. They acquire \emph{face} from image--text data, where a face is mostly a person's, so a registered face arrives already read as human. In CLIP's text space, \emph{a face} indeed lies closest to \emph{a human face}, more clearly for ViT-L/14, which also answers Human more often (Supp.~S12), while the vision-only ViT, which matches regions to annotated examples rather than to words, shows no such collapse. A model asked what kind of face it sees may thus, in effect, be answering whether a person is there, and keeping the two apart calls for training signals in which a face need not belong to a person.

\paragraph{Semantic attribution is not evidence of visual selectivity}
Face detectors almost never accept these regions, while several models name them Human, yet a Human label by itself does not show that the face-like arrangement was used. Our controls expose this separation: Human answers can stay as frequent once the arrangement is removed (Qwen2-VL) or become more frequent (CLIP), while a model that rarely answers Human can still reserve it for the face-like regions (SmolVLM). A claim that a model sees faces in things therefore needs evidence that its Human answers fall once the face-like arrangement is gone, together with the description under which they do.

\paragraph{Task language shapes the judgment being measured}
Category descriptions help define the visual judgment being made. Extending Human from ``a human face'' to ``human-like face patterns'' changes which appearances the category admits, so a change in Human rate need not reflect a change in visual sensitivity. A broader category can explain why more face-like regions are called Human, but not why patches with no structure are: under Face-explicit, LLaVA-1.5 calls every mean-color patch Human, and Qwen2-VL does so once the categories are glossed as kinds of face. The competing texts matter as much: under CLIP's four-text ensemble, control inputs move from Human to Other although the images are unchanged. People who see a face in an outlet still judge that no person is there; a model must express both through a single category word, whose wording and rivals shape both.

\paragraph{Less visual evidence need not yield less specific claims}
Pareidolia offers a resemblance that supports more than one reading; mean-color patches remove even that, yet Qwen2-VL, and LLaVA-1.5 under Face-explicit, call every patch Human, and Qwen2-VL describes a person in every patch tested, so this is not only a product of forced choice. Absent evidence need not make a response more cautious; a model can instead supply content that the image does not support. For Qwen2-VL, the text-only test locates the source in language priors, on which VLMs are known to fall back when visual evidence is weak~\cite{leng2024vcd,favero2024m3id,lee2025vlind}. Evaluation on ordinary photographs, where a person is usually present, would not reveal such false positives. Recognizing people when they are there is one side of the capability; withholding person claims when the image gives no support is the other.

\paragraph{Different routes to the same label}
These behaviors arise from different decision rules: ViT assigns the nearest visual prototype, CLIP the best-matching category text, and the generative models produce a word conditioned on both image and instruction. Neither interface type nor size settles the outcome: the same model changes its selectivity when only the description changes, and SmolVLM and Qwen2-VL, both about 2B parameters, behave in opposite ways.

\paragraph{Implications for evaluation}
Three practices follow. Human rates on ambiguous faces should be reported alongside rates on matched controls, including a structureless input that shows what a model says when nothing is there. Category descriptions should be treated as experimental variables, not chosen by how many Human answers they produce: here, the description that maximized Human answers removed selectivity. Where the target category carries consequences, as in person detection for privacy or safety, a description that announces or widens the category can inflate detections without improving their grounding.

\paragraph{Scope and outlook}
Because the pareidolic selectivity test needs only a model's answers, it applies to any vision model, open or closed, without access to its internals, and extends unchanged to larger or differently trained models. Its conclusions rest on converging controls, and for CLIP, phase-scrambled and blurred regions reproduce the scrambled result. Pairing the test with analyses of internal representations is a natural next step to locate where Human answers form, and the same design carries over to other pareidolia datasets and to any category whose visual evidence is ambiguous. The test can thus serve as a check before a vision model is trusted to report whether a person is present, and as a common ground for comparing how models and people resolve ambiguous faces.

\section{CONCLUSION}

Using face pareidolia as a probe, we asked whether vision models call face-like objects Human because of their face-like arrangement. Models do respond to illusory faces, but for several of them a face is a human face by default: where people read animals or cartoons, they answer Human, even when the answer follows the face. Often, however, the answer does not follow the face: in four of six models, Human answers persist or grow when the arrangement is removed, and changing only the description can remove or create a preference for the face. A Human label is therefore not evidence of face selectivity; it must be read against controls and the category description. Pareidolia exposes this gap by separating resemblance from identity, and the same test applies wherever visual evidence is ambiguous.

{\small
\bibliographystyle{ieee}
\bibliography{egbib}
}

\clearpage
\renewcommand{\topfraction}{0.9}
\renewcommand{\bottomfraction}{0.8}
\renewcommand{\textfraction}{0.07}
\renewcommand{\floatpagefraction}{0.75}
\renewcommand{\dbltopfraction}{0.9}
\renewcommand{\dblfloatpagefraction}{0.75}
\setcounter{topnumber}{4}
\setcounter{bottomnumber}{3}
\setcounter{totalnumber}{8}
\setcounter{dbltopnumber}{3}
\setcounter{section}{0}
\setcounter{subsection}{0}
\setcounter{table}{0}
\setcounter{figure}{0}
\setcounter{equation}{0}
\renewcommand{\thesection}{S\arabic{section}}
\renewcommand{\thetable}{S\arabic{table}}
\renewcommand{\thefigure}{S\arabic{figure}}
\renewcommand{\theequation}{S\arabic{equation}}

\twocolumn[{%
\begin{center}
{\LARGE \bf Supplementary Material}\\[4pt]
{\large Faces in Things, Faces in Nothing: Probing Face Selectivity in Vision Models with Pareidolia}
\end{center}
\vspace{1em}
}]

This supplement documents the data preparation, model configurations, category descriptions, and control construction used in the main paper, and reports the full results and additional analyses referred to there. Section and figure numbers prefixed with ``S'' refer to this document; all others refer to the main paper. Table~\ref{tab:s_guide} lists the questions a reader may have and where each is answered.

\begin{table}[!htbp]
\centering
\caption{Reader's guide: where each question is answered}
\label{tab:s_guide}
\setlength{\tabcolsep}{3pt}
{\footnotesize
\begin{tabular}{@{}p{6.75cm}l@{}}
\toprule
Question & Section \\
\midrule
\multicolumn{2}{@{}l}{\textit{Setup}}\\
How are regions selected and labeled? & S1 \\
Which checkpoints and settings are used? & S2 \\
What exactly does each description say? & S3 \\
How are the controls built, and could they bias the result? & S4 \\
How are free-text answers parsed? & S5 \\
\midrule
\multicolumn{2}{@{}l}{\textit{Do models respond to illusory faces, and as people do?}}\\
Do face detectors respond to these regions? & S8 \\
Does Human follow how people read the face? & S9 \\
Does \emph{face} mean a human face in CLIP's language? & S12 \\
\midrule
\multicolumn{2}{@{}l}{\textit{Do Human answers follow the face?}}\\
Full Human rates and $\Delta_c$ with CIs, all models and descriptions & S6 \\
Do models call real faces Human? & S7 \\
Is the Human answer only a product of forced choice? & S7 \\
Does Qwen2-VL answer Human without any image? & S7 \\
Does CLIP's scrambled result hold without tile edges? & S4, S11 \\
Does it hold on all $4{,}536$ regions, not just $400$? & S11 \\
\midrule
\multicolumn{2}{@{}l}{\textit{What does the description change?}}\\
How many answers change, and to which categories? & S10 \\
Where do CLIP's control inputs go under each text set? & S11 \\
\bottomrule
\end{tabular}
}
\end{table}

\section{Data and Labels}
\label{sec:s_data}

We use the public FacesInThings release~\cite{hamilton2024seeing} ($4{,}536$ decodable images). For every image, we crop the box flagged as primary, or the first box when none is flagged, without padding; all interfaces receive this same crop. Each region takes one of five categories, Human, Animal, Cartoon, Alien, or Other, from the annotators' reading of the face; the $14$ regions without this annotation are excluded from label-conditioned analyses.

\section{Model Configurations}
\label{sec:s_models}

All models use publicly released weights without fine-tuning. Experiments ran on an Apple M4 MacBook Air (32\,GB unified memory) with Python~3.9, PyTorch~2.8.0, transformers~4.57.6, open\_clip~3.2.0, timm~1.0.24, insightface~0.7.3, and ultralytics (YOLOv8). Table~\ref{tab:s_models} summarizes the configurations. Generative models run in float16 on Apple MPS; a float32 check is reported in Sec.~\ref{sec:s_sanity}.

\begin{table*}[t]
\centering
\caption{Model configurations}
\label{tab:s_models}
{\footnotesize
\begin{tabular}{p{2.6cm}p{5.4cm}p{8.4cm}}
\toprule
Interface & Weights & Inference \\
\midrule
ViT-B/16 & timm \texttt{vit\_base\_patch16\_224} (default pretrained weights) & ImageNet logits as features; $\ell_2$-normalized; nearest leave-one-out class prototype by cosine similarity (Sec.~\ref{sec:s_vit}). \\
CLIP ViT-B/32, ViT-L/14 & open\_clip \texttt{ViT-B-32}, \texttt{ViT-L-14}, OpenAI weights, \texttt{force\_quick\_gelu=True} & One $\ell_2$-normalized text embedding per category (mean of normalized embeddings for multi-text descriptions); argmax of cosine similarity. \\
LLaVA-1.5-7B & \texttt{llava-hf/llava-1.5-7b-hf} & Float16 on MPS; user prompt with the crop; no system prompt; greedy decoding; at most $16$ new tokens. \\
Qwen2-VL-2B & \texttt{Qwen/Qwen2-VL-2B-Instruct} & As above; image passed through the chat template and processor. \\
SmolVLM & \texttt{HuggingFaceTB/SmolVLM-Instruct} & As above. \\
RetinaFace & insightface \texttt{buffalo\_l} pack (RetinaFace-10GF detector), input size $640$ & Run on each annotated box cropped with $10\%$ padding; any face detection counts as a response. \\
YOLOv8 & ultralytics \texttt{yolov8n.pt} & As above; any \emph{person} detection counts as a response. \\
\bottomrule
\end{tabular}
}
\end{table*}

\subsection{Leave-one-out ViT prototypes}
\label{sec:s_vit}
Prototypes are built from the intact primary crops of all $4{,}522$ labeled regions. For a region $i$ with annotation $c_i$, the prototype of category $c$ is the normalized mean feature of all crops annotated $c$, excluding crop $i$ itself. The same exclusion is applied when region $i$'s background, scrambled, or mean-color control is classified, so no evaluated image ever contributes to the prototype it is compared with. Controls never enter any prototype.

\section{Category Descriptions}
\label{sec:s_prompts}

\subsection{Prompts for generative models}
The three descriptions are identical across LLaVA, Qwen2-VL, and SmolVLM. The Defined prompt reads (line breaks added for layout):
{\scriptsize
\begin{verbatim}
You will see a small cropped region of a larger
image. Classify the MAIN pattern into exactly ONE
category:
- human : a human face
- animal : an animal face
- cartoon : a cartoon / emoji / animated character face
- alien : an alien / monster / unrealistic creature face
- other : none of the above

Important:
- If uncertain, choose the SINGLE best category.
- Do NOT explain your reasoning.

Output ONLY ONE WORD in English, lowercase, from:
`human`, `animal`, `cartoon`, `alien`, `other`.
\end{verbatim}
}
\emph{Bare} replaces the category block by ``Classify the MAIN pattern into exactly ONE of:'' followed by the five names without glosses. \emph{Face-explicit} inserts ``This crop may contain a face-like pattern.'' after the first sentence and changes two glosses to ``human : a human face or human-like face pattern'' and ``animal : an animal face or animal-like face pattern''. All other text is unchanged.

\subsection{CLIP category texts}
Table~\ref{tab:s_cliptexts} lists the texts. Bare, Defined, and Face-explicit use exactly one text per category, matching the generative glosses. The ensemble uses four texts per category and describes Other through concrete non-face content. The face-only set, used for the CLIP analyses on all regions (Sec.~\ref{sec:s_clip}) and in Fig.~7 of the main paper, uses one short text per category and describes Other only by negation.

\begin{table*}[t]
\centering
\caption{CLIP category texts. Multi-text entries are averaged after normalization}
\label{tab:s_cliptexts}
{\footnotesize
\begin{tabular}{lp{1.1cm}p{2.7cm}p{2.9cm}p{1.6cm}p{5.8cm}}
\toprule
Category & Bare & Defined & Face-explicit & Face-only & Ensemble (four texts) \\
\midrule
Human & human & a human face & a human face or human-like face pattern & a human face & a human face; a human-like face illusion (pareidolia); two eyes and a mouth forming a face; a human face-like pattern in objects \\
Animal & animal & an animal face & an animal face or animal-like face pattern & an animal face & an animal face; an animal-like face illusion; a dog or cat-like face in objects; an animal face-like pattern \\
Cartoon & cartoon & a cartoon / emoji / animated character face & (as Defined) & a cartoon face & a cartoon face; an animated character face; a smiling emoji-like face; a stylized cartoon face pattern \\
Alien & alien & an alien / monster / unrealistic creature face & (as Defined) & an alien face & an alien face; an extraterrestrial creature face; a non-human creature face; a strange alien-like face pattern \\
Other & other & none of the above & (as Defined) & not a face & an object pattern with no face; a random object without any face; a texture or pattern that is not a face; household objects without a face \\
\bottomrule
\end{tabular}
}
\end{table*}

\section{Controls}
\label{sec:s_controls}

\paragraph{Same-image background}
For an image of size $W{\times}H$, the background crop is a square of side $s=\max(96, \lfloor\min(W,H)/3\rfloor)$. Every annotated box in the image (not only the primary one) is expanded by $2$ pixels, and up to $200$ candidate positions are drawn uniformly with a generator seeded from the file name. The first candidate that overlaps no expanded box is kept; if none qualifies, the image is skipped. Images annotated as non-human are visited in a fixed random order (seed $42$) until $400$ backgrounds are found; $99$ images were skipped along the way. The controlled subset thus consists of images in which the face-like region leaves room for a background of the same scene.

\begin{figure*}[t]
\centering
\includegraphics[width=0.92\linewidth]{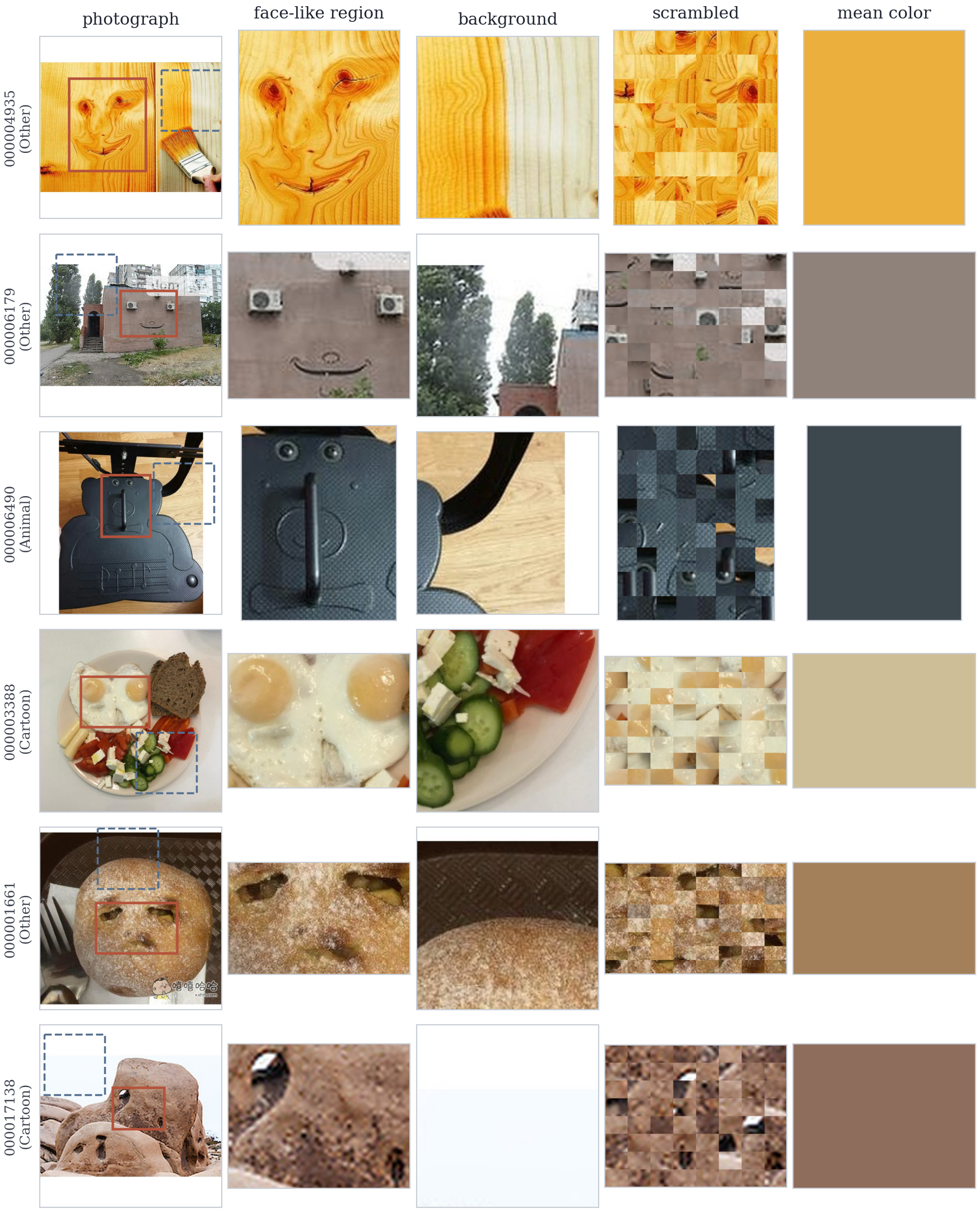}
\caption{The four inputs for six images of the controlled subset. Left: photograph with the face-like region (solid) and the background crop (dashed); the annotated reading is given in parentheses. The last row shows a background that falls largely on white padding.}
\label{fig:s_examples}
\end{figure*}

\paragraph{Scrambled and mean-color versions}
The scrambled control divides the region into an $8{\times}8$ grid and permutes the $64$ tiles; remainder rows or columns that do not fill a complete tile are dropped. The permutation is drawn from a generator seeded with $(42, \mathrm{adler32}(\text{file name}))$, so every model sees the identical scrambled image. The mean-color control fills the region with its mean RGB value. Fig.~\ref{fig:s_examples} shows examples.

\paragraph{Controls without tile edges}
To test whether CLIP's high Human rate on scrambled regions depends on tile boundaries, we add three manipulations of the same $400$ regions: a \emph{phase-scrambled} version (amplitude spectrum of each channel kept, one shared random phase field for all channels, mean and standard deviation of each channel restored), a \emph{blurred} version (Gaussian blur with radius $\max(12, \lfloor\min(w,h)/6\rfloor)$), and the intact region overlaid with an $8{\times}8$ grid of lines (tile edges without shuffling). Results are in Sec.~\ref{sec:s_clip}.

\paragraph{Near-blank backgrounds}
Some FacesInThings images are padded to a square canvas, and a background crop can fall partly on the padding. Of the $400$ backgrounds, $40$ ($10.0\%$) consist of more than $50\%$ near-white pixels (all channels above $245$), and $7$ ($1.8\%$) are near-uniform (mean per-channel standard deviation below $8$). Excluding the $40$ mostly white backgrounds leaves every region-versus-background difference within two percentage points. Background answers are similar for plain and textured scenes: splitting the backgrounds at their median pixel standard deviation, Qwen2-VL calls the more textured half Human more often than the plainer half under Defined ($56.5\%$ against $46.5\%$), and the corresponding rates are $13.5\%$ against $11.0\%$ for LLaVA-1.5, $25.5\%$ against $31.0\%$ for CLIP ViT-B/32, and $49.0\%$ against $52.0\%$ for CLIP ViT-L/14.

\paragraph{People in background crops}
Background crops are checked for real people. Run on the $400$ background crops, RetinaFace returns a face for one and YOLOv8 a person for $26$. Removing the $26$ crops in which YOLOv8 detects a person changes no background Human rate by more than $1.7$ percentage points.

\begin{table*}[t]
\centering
\caption{Human rate (\%) on the four inputs of the $400$ controlled images and pareidolic selectivity $\Delta_c$ (percentage points) with $95\%$ bootstrap intervals. B: same-image background; S: scrambled region; M: mean-color region. $^\ddagger$SmolVLM Bare background: $397$ valid answers; Qwen2-VL Bare scrambled: $399$}
\label{tab:s_matrix}
\setlength{\tabcolsep}{2.5pt}
{\scriptsize
\begin{tabular}{llrrrrccc}
\toprule
Interface & Description & Region & B & S & M & $\Delta_{\mathrm{B}}$ & $\Delta_{\mathrm{S}}$ & $\Delta_{\mathrm{M}}$ \\
\midrule
LLaVA-1.5 & Bare & $11.0$ & $16.8$ & $5.8$ & $12.0$ & $-5.8$ [$-10.5$, $-1.5$] & $+5.2$ [$2.0$, $8.8$] & $-1.0$ [$-5.2$, $3.5$] \\
 & Defined & $59.0$ & $12.2$ & $25.0$ & $0.0$ & $+46.8$ [$41.2$, $52.2$] & $+34.0$ [$27.8$, $40.2$] & $+59.0$ [$54.5$, $63.7$] \\
 & Face-explicit & $96.2$ & $99.5$ & $100.0$ & $100.0$ & $-3.2$ [$-5.2$, $-1.2$] & $-3.8$ [$-5.8$, $-2.0$] & $-3.8$ [$-5.8$, $-2.0$] \\
\addlinespace
Qwen2-VL & Bare & $8.8$ & $8.5$ & $10.5$ & $0.0$ & $+0.2$ [$-3.5$, $4.5$] & $-2.0$ [$-6.0$, $1.8$] & $+8.8$ [$6.2$, $11.5$] \\
 & Defined & $51.2$ & $51.5$ & $59.0$ & $100.0$ & $-0.2$ [$-7.2$, $7.0$] & $-7.8$ [$-14.3$, $-1.2$] & $-48.8$ [$-53.8$, $-43.8$] \\
 & Face-explicit & $80.5$ & $86.0$ & $98.8$ & $100.0$ & $-5.5$ [$-10.5$, $-0.8$] & $-18.2$ [$-22.2$, $-14.2$] & $-19.5$ [$-23.5$, $-15.5$] \\
\addlinespace
SmolVLM & Bare & $4.0$ & $1.5$ & $0.2$ & $0.0$ & $+2.5$ [$0.3$, $4.8$] & $+3.8$ [$2.0$, $5.8$] & $+4.0$ [$2.2$, $6.0$] \\
 & Defined & $4.8$ & $0.2$ & $0.2$ & $0.0$ & $+4.5$ [$2.5$, $6.8$] & $+4.5$ [$2.5$, $6.8$] & $+4.8$ [$2.8$, $7.0$] \\
 & Face-explicit & $6.0$ & $1.2$ & $0.5$ & $0.0$ & $+4.8$ [$2.2$, $7.2$] & $+5.5$ [$3.2$, $8.0$] & $+6.0$ [$3.8$, $8.2$] \\
\addlinespace
CLIP ViT-B/32 & Bare & $22.5$ & $14.5$ & $41.2$ & $43.2$ & $+8.0$ [$2.5$, $13.8$] & $-18.8$ [$-24.8$, $-13.0$] & $-20.8$ [$-27.3$, $-14.8$] \\
 & Defined & $18.8$ & $28.2$ & $86.0$ & $31.0$ & $-9.5$ [$-14.8$, $-4.2$] & $-67.2$ [$-72.0$, $-62.5$] & $-12.2$ [$-18.2$, $-6.0$] \\
 & Face-explicit & $7.8$ & $14.5$ & $74.8$ & $3.5$ & $-6.8$ [$-10.8$, $-2.8$] & $-67.0$ [$-71.5$, $-62.5$] & $+4.2$ [$0.8$, $7.5$] \\
 & Ensemble & $36.2$ & $29.8$ & $22.0$ & $0.0$ & $+6.5$ [$0.5$, $13.0$] & $+14.2$ [$8.2$, $20.2$] & $+36.2$ [$31.8$, $41.0$] \\
\addlinespace
CLIP ViT-L/14 & Bare & $58.8$ & $24.5$ & $67.8$ & $87.8$ & $+34.2$ [$28.0$, $40.2$] & $-9.0$ [$-15.0$, $-3.2$] & $-29.0$ [$-34.8$, $-23.0$] \\
 & Defined & $43.8$ & $50.5$ & $86.8$ & $86.8$ & $-6.8$ [$-13.2$, $-0.5$] & $-43.0$ [$-48.2$, $-37.5$] & $-43.0$ [$-48.8$, $-37.0$] \\
 & Face-explicit & $16.5$ & $10.5$ & $81.5$ & $27.8$ & $+6.0$ [$1.2$, $10.5$] & $-65.0$ [$-70.0$, $-59.8$] & $-11.2$ [$-17.2$, $-5.8$] \\
 & Ensemble & $35.5$ & $19.5$ & $19.8$ & $0.0$ & $+16.0$ [$10.2$, $22.0$] & $+15.8$ [$10.2$, $21.2$] & $+35.5$ [$31.0$, $40.2$] \\
\addlinespace
ViT-B/16 & -- & $5.0$ & $9.0$ & $8.5$ & $0.5$ & $-4.0$ [$-7.2$, $-0.8$] & $-3.5$ [$-6.5$, $-0.5$] & $+4.5$ [$2.2$, $7.0$] \\
\bottomrule
\end{tabular}
}
\end{table*}

\section{Answer Parsing and Validation}
\label{sec:s_parsing}

The strict parser accepts an answer only if it is a single line that, after lowercasing and removing surrounding quotes and punctuation, is exactly one of the five category names; anything else, including longer phrases, multiple lines, or negations such as ``not human'', is marked invalid rather than mapped to a category. The main-paper predictions on the $3{,}408$ regions use an earlier parser that maps unparseable answers to Other. Regenerating answers for $80$ randomly sampled regions under all three models and descriptions ($720$ answers) shows that the two parsers agree on all $719$ answers valid under both; the remaining answer (SmolVLM, Bare) is unparseable and counted as Other.

All control runs (backgrounds, scrambled, and mean-color inputs) use the strict parser and exclude invalid answers from both numerator and denominator. Invalid answers are rare: SmolVLM returned three background answers outside the categories under Bare (\texttt{watermelon}, \texttt{apple}, \texttt{orange}), and Qwen2-VL returned one unparseable answer for a scrambled region under Bare (\texttt{wood}). All other conditions are complete ($n{=}400$).

\section{Full Control Results}
\label{sec:s_matrix}

Table~\ref{tab:s_matrix} reports the Human rate of every interface and description on the four inputs of the $400$ controlled images, together with the pareidolic selectivity $\Delta_c = p_{r,d} - p_{c,d}$ for each control $c$ and its image-paired percentile bootstrap $95\%$ interval ($2{,}000$ resamples).

\section{Real Faces, Open-ended Descriptions, and a Text-only Test}
\label{sec:s_sanity}

\paragraph{Real faces}
As a positive reference, we use $500$ face photographs drawn at random (seed $42$) from Labeled Faces in the Wild~\cite{huang2008lfw}, in the tightly cropped format distributed with scikit-learn (about $125{\times}94$ pixels). The generative models see the first $400$ of them, CLIP ViT-B/32 all $500$. Table~\ref{tab:s_lfw} reports Human rates. All three generative models call nearly every real face Human; SmolVLM's few other answers are all Other. CLIP ViT-B/32 calls $70.6\%$ Human with one text per category, assigning most of the rest to Alien; with the face-only set (``a human face'' $\ldots$ ``not a face'') it calls $97.2\%$ Human.

\begin{table}[!htbp]
\centering
\caption{Human rate (\%) on real faces (LFW)}
\label{tab:s_lfw}
\setlength{\tabcolsep}{4pt}
{\footnotesize
\begin{tabular}{lrrr}
\toprule
Model ($n$) & Bare & Defined & Face-explicit \\
\midrule
LLaVA-1.5 ($400$) & $99.8$ & $100.0$ & $100.0$ \\
Qwen2-VL ($400$) & $100.0$ & $100.0$ & $100.0$ \\
SmolVLM ($400$) & $98.5$ & $93.0$ & $94.8$ \\
CLIP ViT-B/32 ($500$) & -- & $70.6$ & -- \\
\quad ensemble / face-only & \multicolumn{3}{c}{$69.8$ / $97.2$} \\
\bottomrule
\end{tabular}
}
\end{table}

\paragraph{Open-ended descriptions}
To see what the generative models report without the category task, we asked each to ``Describe this image in a few words.'' for $20$ mean-color and $20$ scrambled inputs sampled from the controlled subset (seed $42$; greedy decoding; at most $48$ new tokens). Table~\ref{tab:s_open} summarizes keyword tags and quotes typical answers. On mean-color patches, LLaVA-1.5 describes a wall in a given color and SmolVLM a solid color; Qwen2-VL describes a person in all $20$ cases, with only four distinct answers and no mention of the color. On scrambled regions all three models describe mosaics, collages, patterns, or objects; LLaVA-1.5 mentions a face in two of $20$.

\begin{table*}[t]
\centering
\caption{Open-ended descriptions of $20$ mean-color and $20$ scrambled inputs: fraction of answers mentioning a color or blank surface, noise or texture, and a face or person (keyword tags), with typical answers}
\label{tab:s_open}
\setlength{\tabcolsep}{3pt}
{\footnotesize
\begin{tabular}{llrrrp{8.4cm}}
\toprule
Model & Input & color/blank & texture & face/person & Typical answer \\
\midrule
LLaVA-1.5 & mean color & $0.90$ & $0.45$ & $0.00$ & ``The image is a close-up of a gray wall, possibly a concrete wall.'' \\
 & scrambled & $0.10$ & $0.35$ & $0.10$ & ``A collage of various food items, including eggs, tomatoes, and bacon.'' \\
Qwen2-VL & mean color & $0.00$ & $0.00$ & $1.00$ & ``The image shows a person standing in a room with a white wall and a window.'' \\
 & scrambled & $0.35$ & $0.75$ & $0.05$ & ``A mosaic with beige and brown tiles.'' \\
SmolVLM & mean color & $1.00$ & $0.00$ & $0.00$ & ``The image is a solid grey color.'' \\
 & scrambled & $0.30$ & $0.50$ & $0.00$ & ``This image is a collage of many small squares.'' \\
\bottomrule
\end{tabular}
}
\end{table*}

\paragraph{Numerical precision}
All generative runs use float16 on Apple MPS. Rerunning Qwen2-VL in float32 on CPU ($10$ each of face-like regions, scrambled regions, mean-color patches, and real faces; Defined, Face-explicit, and the open-ended prompt) reproduces $118$ of $120$ answers character for character and every category label: Qwen2-VL again calls every mean-color patch Human and describes a person in each.

\paragraph{Text-only test for Qwen2-VL}
To test whether Qwen2-VL's answers on structureless inputs depend on the image, we gave it the paper's prompts with no image (a text-only user turn), and with solid white and black $224{\times}224$ images. To separate a preference for the human category from a preference for a list position, the Defined and Bare prompts were also run with human moved from the first to the last place, both in the category list and in the list of allowed outputs; all other text was unchanged. Decoding is greedy, so each cell is a single deterministic answer (Table~\ref{tab:s_prior}). Under Defined, Qwen2-VL answers human without any image, and still does so when human is listed last, so the answer reflects the face glosses rather than list position. Under Bare it answers with whichever category is listed last (other, or human when human is moved to the end), so its Bare answers on structureless inputs reflect list position. Asked to describe a missing image, it produces the same person description it gives for mean-color patches (``a person standing in a room with a white wall and a window''); white and black images also yield person descriptions. For this model, a featureless patch is thus treated like no image.

\begin{table}[!htbp]
\centering
\caption{Qwen2-VL answers without an image and on solid images (one greedy answer per cell). ``first'' and ``last'' give the position of human in the category list}
\label{tab:s_prior}
\setlength{\tabcolsep}{3pt}
{\footnotesize
\begin{tabular}{lccccl}
\toprule
Input & \multicolumn{2}{c}{Defined} & \multicolumn{2}{c}{Bare} & Description \\
 & first & last & first & last & \\
\midrule
No image & human & human & other & human & ``a person standing\ldots'' \\
White & human & -- & other & -- & ``a person is standing\ldots'' \\
Black & human & -- & other & -- & ``a person standing\ldots'' \\
\bottomrule
\end{tabular}
}
\end{table}

\section{Face Detection Reference}
\label{sec:s_detectors}

Detectors are run on every annotated region cropped with $10\%$ padding; any face (RetinaFace) or person (YOLOv8) detection counts as a response. Table~\ref{tab:s_det_primary} reports response rates on the $4{,}522$ labeled primary regions. RetinaFace responds to $51$ of them ($1.1\%$). Its responses are more frequent for regions annotated as human-like ($28/1{,}114$, $2.5\%$) than for regions annotated as non-human ($23/3{,}408$, $0.7\%$; two-proportion $z=5.1$, $p<10^{-6}$), and for faces annotated as easy to spot ($41/2{,}726$, $1.5\%$) than for hard ones ($2/813$, $0.2\%$; $z=2.9$, $p=0.004$). YOLOv8's person responses are more frequent ($8.6\%$) but do not decrease with difficulty, and on the $400$ controlled images they are as frequent for background crops ($6.5\%$) as for the face-like regions ($6.25\%$); RetinaFace responds to one region and one background there.

\begin{table}[!htbp]
\centering
\caption{Detector response rate (\%) on the $4{,}522$ labeled primary regions, by annotated category and difficulty}
\label{tab:s_det_primary}
\setlength{\tabcolsep}{3.5pt}
{\footnotesize
\begin{tabular}{lrrrrrr}
\toprule
 & Human & Animal & Cartoon & Alien & Other & non-human \\
\midrule
RetinaFace & $2.5$ & $1.2$ & $0.1$ & $0.3$ & $0.7$ & $0.7$ \\
YOLOv8 & $12.4$ & $9.0$ & $4.9$ & $10.1$ & $6.7$ & $7.3$ \\
\midrule
 & Easy & Medium & Hard & & & all \\
\midrule
RetinaFace & $1.5$ & $0.9$ & $0.2$ & & & $1.1$ \\
YOLOv8 & $7.7$ & $10.7$ & $9.2$ & & & $8.6$ \\
\bottomrule
\end{tabular}
}
\end{table}

\section{Human Answers by Annotated Category}
\label{sec:s_bycat}

Table~\ref{tab:s_bycat} gives the Human rate under Defined for each annotated category on all labeled regions. Every model answers Human more often for regions annotated as human-like than for the rest. Models that can answer Other mostly treat the regions as faces: LLaVA-1.5, Qwen2-VL, and ViT assign one of the four kinds of face to $81.3\%$, $88.0\%$, and $85.4\%$ of all labeled regions, whereas SmolVLM does so for $13.6\%$. In the three models with high Human rates, faces annotated as alien are called Human most often and faces annotated as animal least often. The difference between human-like and non-human regions stays modest even as the overall rate rises tenfold: $11.5$ points for ViT, $16.3$ for LLaVA-1.5, $24.1$ for Qwen2-VL, and $23.1$ for CLIP ViT-L/14.

\begin{table}[!htbp]
\centering
\caption{Human rate (\%) by annotated category under Defined ($4{,}522$ labeled regions)}
\label{tab:s_bycat}
\setlength{\tabcolsep}{3pt}
{\footnotesize
\begin{tabular}{lrrrrrr}
\toprule
Model & Human & Animal & Cartoon & Alien & Other & non-human \\
\midrule
ViT-B/16 & $16.1$ & $6.4$ & $2.6$ & $7.2$ & $3.6$ & $4.6$ \\
SmolVLM & $18.3$ & $6.0$ & $5.4$ & $2.1$ & $4.5$ & $4.9$ \\
CLIP ViT-B/32 & $27.1$ & $9.0$ & $20.7$ & $16.4$ & $16.0$ & $14.9$ \\
CLIP ViT-L/14 & $67.8$ & $34.1$ & $44.4$ & $55.8$ & $50.1$ & $44.7$ \\
Qwen2-VL & $74.8$ & $45.8$ & $49.3$ & $65.4$ & $51.6$ & $50.7$ \\
LLaVA-1.5 & $72.4$ & $44.6$ & $53.1$ & $70.4$ & $62.9$ & $56.1$ \\
\bottomrule
\end{tabular}
}
\end{table}

\section{Where Changed Answers Come From}
\label{sec:s_transitions}

\begin{figure}[!htbp]
\centering
\includegraphics[width=0.95\linewidth]{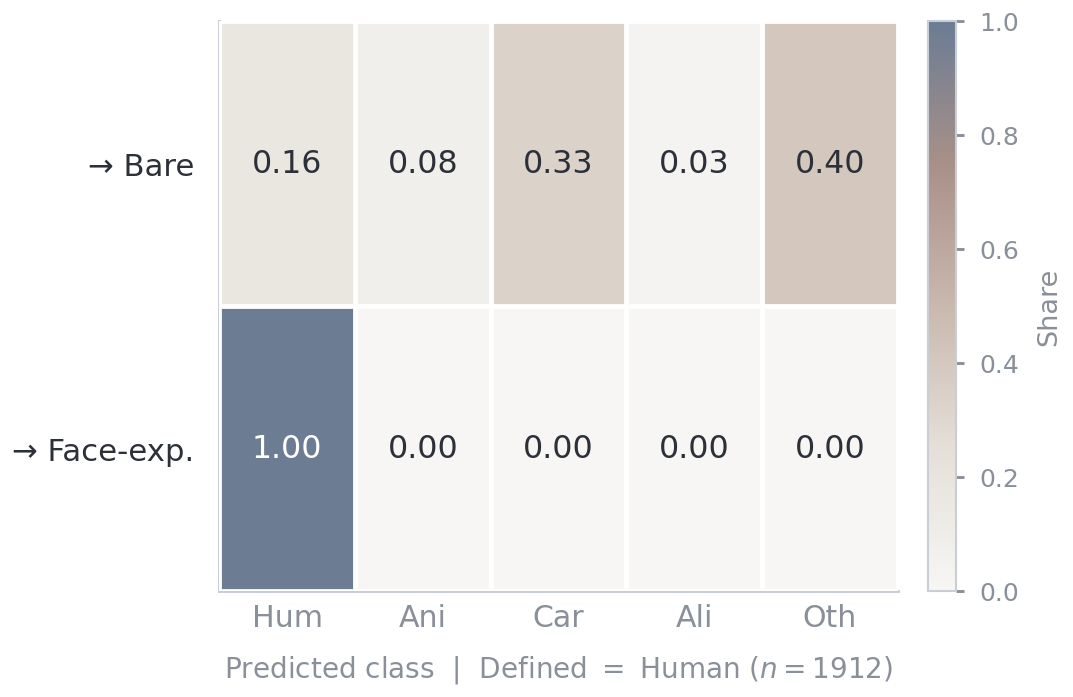}
\caption{Destinations under Bare and Face-explicit of the $1{,}912$ regions that LLaVA-1.5 labels Human under Defined.}
\label{fig:s_transitions}
\end{figure}

Table~\ref{tab:s_switch} reports how often the predicted category of a region changes between Defined and another description. For LLaVA-1.5, all $1{,}912$ regions labeled Human under Defined remain Human under Face-explicit, and $92.9\%$ of the $1{,}409$ newly Human regions come from Other or Cartoon; under Bare, most of the $1{,}912$ move to Other or Cartoon (Fig.~\ref{fig:s_transitions}). Qwen2-VL behaves alike, keeping $1{,}716$ of its $1{,}728$ Defined-Human regions under Face-explicit. For SmolVLM, most changes do not involve Human.

\begin{table}[!htbp]
\centering
\caption{Label-switch rate (\%) relative to Defined. Generative models: $3{,}408$ regions annotated as non-human (count in parentheses). CLIP: the $400$ controlled regions}
\label{tab:s_switch}
\setlength{\tabcolsep}{3pt}
{\footnotesize
\begin{tabular}{lccc}
\toprule
Model & vs.\ Bare & vs.\ Face-explicit & vs.\ ensemble \\
\midrule
LLaVA-1.5 & $51.5$ ($1{,}756$) & $41.4$ ($1{,}410$) & -- \\
Qwen2-VL & $47.1$ ($1{,}605$) & $30.8$ ($1{,}051$) & -- \\
SmolVLM & $19.5$ ($665$) & $5.1$ ($174$) & -- \\
CLIP ViT-B/32 & $63.7$ & $22.8$ & $50.5$ \\
CLIP ViT-L/14 & $51.5$ & $36.5$ & $38.2$ \\
\bottomrule
\end{tabular}
}
\end{table}

\section{CLIP: All Regions, Label Destinations, and Controls without Tile Edges}
\label{sec:s_clip}

\paragraph{All regions}
CLIP ViT-B/32 was also run on the scrambled and mean-color versions of all $4{,}536$ regions and on the $500$ LFW faces, under three category-text sets (Table~\ref{tab:s_clip_full}). The pattern matches the controlled subset: with one text per category, scrambled regions are called Human far more often than intact ones and more often than real faces; with the four-text ensemble, Human is most frequent for intact regions and absent for mean-color patches. A single-text set whose Other reads ``not a face'' sends $86\%$ of mean-color patches to Human.

\begin{table}[!htbp]
\centering
\caption{CLIP ViT-B/32 Human rate (\%) on all $4{,}536$ regions and their controls, and on $500$ real faces}
\label{tab:s_clip_full}
\setlength{\tabcolsep}{3pt}
{\footnotesize
\begin{tabular}{lrrrr}
\toprule
Category texts & Region & Scrambled & Mean color & Real faces \\
\midrule
Defined (one text) & $17.9$ & $86.4$ & $30.4$ & $70.6$ \\
Face-only (``not a face'') & $13.4$ & $74.7$ & $86.2$ & $97.2$ \\
Ensemble (four texts) & $43.4$ & $22.6$ & $0.0$ & $69.8$ \\
\bottomrule
\end{tabular}
}
\end{table}

\paragraph{Label destinations}
Fig.~7 of the main paper shows where the scrambled and mean-color inputs go: with ``none of the above'' or ``not a face'' as Other, they fall to the face texts, mostly to Human; with concrete non-face descriptions of Other, most scrambled inputs and nearly all mean-color patches go to Other.

\paragraph{Controls without tile edges}
Table~\ref{tab:s_clip_artifact} reports both CLIP models on the $400$ controlled regions under Defined and the ensemble. With one text per category, phase-scrambled and blurred regions, which contain no tile boundaries, are called Human at rates similar to tile-scrambled ones ($84$--$98\%$ versus $86$--$87\%$), and almost none is assigned to Other; overlaying grid lines on the intact region also raises Human answers. With the ensemble, most blurred and tile-scrambled inputs go to Other. Under the ensemble, CLIP ViT-L/14 still calls $58\%$ of phase-scrambled regions Human, more than the intact regions. Predictions on intact and tile-scrambled regions in this run agree with the main run on every image.

\begin{table}[!htbp]
\centering
\caption{CLIP Human rate (\%) on the $400$ controlled regions and on controls without tile edges; percentage assigned to Other in parentheses}
\label{tab:s_clip_artifact}
\setlength{\tabcolsep}{2.5pt}
{\footnotesize
\begin{tabular}{llrrrrr}
\toprule
Model & Texts & Region & Tiles & Phase & Blur & Grid lines \\
\midrule
B/32 & Defined & $18.8$ & $86.0$ ($0$) & $96.0$ ($0$) & $83.8$ ($6$) & $71.8$ ($2$) \\
 & Ensemble & $36.2$ & $22.0$ ($63$) & $31.2$ ($57$) & $0.0$ ($100$) & $21.0$ ($44$) \\
L/14 & Defined & $43.8$ & $86.8$ ($1$) & $98.0$ ($0$) & $84.5$ ($3$) & $55.2$ ($4$) \\
 & Ensemble & $35.5$ & $19.8$ ($56$) & $58.0$ ($24$) & $13.5$ ($70$) & $21.5$ ($58$) \\
\bottomrule
\end{tabular}
}
\end{table}

\section{A Face Is a Human Face in CLIP's Text Space}
\label{sec:s_cliptext}

To test whether \emph{face} already means a human face in the models' language, we encode four neutral face phrases with the CLIP text encoders used in the paper (OpenAI weights, QuickGELU) and compute their cosine similarity to the five Defined category texts, exactly as used for classification (Table~\ref{tab:s_cliptext}). For ViT-L/14, all four phrases lie closest to \emph{a human face}; for ViT-B/32, three of four do, and \emph{a face-like pattern} lies marginally closer to \emph{an animal face}. The margin is larger for ViT-L/14 (\emph{a face}: $0.046$ over the next category, against $0.029$ for ViT-B/32), in line with its higher Human rate in the main experiments. With matched short texts for every category (\emph{a cartoon face}, \emph{an alien face}, \emph{an object}), \emph{a face} and \emph{a photo of a face} remain closest to \emph{a human face} in both models; the remaining two phrases occasionally fall closer to Cartoon or Alien.

\begin{table}[!htbp]
\centering
\caption{Cosine similarity between neutral face phrases and the Defined category texts in CLIP's text space. Bold marks the closest category}
\label{tab:s_cliptext}
\setlength{\tabcolsep}{3pt}
{\footnotesize
\begin{tabular}{llccccc}
\toprule
Model & Phrase & Human & Animal & Cartoon & Alien & Other \\
\midrule
ViT-B/32 & a face & $\mathbf{.946}$ & $.917$ & $.875$ & $.893$ & $.845$ \\
 & a photo of a face & $\mathbf{.954}$ & $.926$ & $.847$ & $.876$ & $.837$ \\
 & a face-like pattern & $.879$ & $\mathbf{.882}$ & $.829$ & $.847$ & $.818$ \\
 & a face in an object & $\mathbf{.924}$ & $.911$ & $.839$ & $.873$ & $.810$ \\
\midrule
ViT-L/14 & a face & $\mathbf{.915}$ & $.869$ & $.789$ & $.823$ & $.783$ \\
 & a photo of a face & $\mathbf{.921}$ & $.847$ & $.761$ & $.795$ & $.711$ \\
 & a face-like pattern & $\mathbf{.857}$ & $.841$ & $.783$ & $.762$ & $.715$ \\
 & a face in an object & $\mathbf{.879}$ & $.854$ & $.778$ & $.799$ & $.744$ \\
\bottomrule
\end{tabular}
}
\end{table}

\FloatBarrier

\end{document}